\documentclass[11pt]{article}

\usepackage[]{acl}
\usepackage{xspace}
\usepackage{float}
\usepackage{times}
\usepackage{latexsym}
\usepackage{booktabs}

\usepackage{longtable}
\usepackage{tabularx}
\usepackage{array}
\usepackage{booktabs}
\usepackage[T1]{fontenc}

\usepackage[utf8]{inputenc}

\usepackage{microtype}

\usepackage{inconsolata}
\usepackage{amsmath}

\usepackage{pgfplots}
\pgfplotsset{compat=1.18}
\usepackage{sansmath} 
\tikzset{every mark/.append style={scale=1.1}}
\usepackage{graphicx}

\title{Beyond Consensus: Perspectivist Modeling and Evaluation of Annotator Disagreement in NLP}

\author{Yinuo Xu \\
 University of Michigan \\
  \texttt{yinuoxu@umich.edu} \\\And
  David Jurgens \\
  University of Michigan\\
  \texttt{jurgens@umich.edu} \\}

\usepackage{tabularx}
\usepackage{tabularx}
\usepackage{booktabs}
\usepackage{array}
\usepackage{ragged2e}
\renewcommand{\arraystretch}{1.2}

\newcolumntype{Y}{>{\RaggedRight\arraybackslash}X}
\begin{document}
\maketitle
\begin{abstract}
Annotator disagreement is widespread in NLP, particularly for subjective and ambiguous tasks such as toxicity detection and stance analysis. While early approaches treated disagreement as noise to be removed, recent work increasingly models it as a meaningful signal reflecting variation in interpretation and perspective. This survey provides a unified view of disagreement-aware NLP methods. We first present a domain-agnostic taxonomy of the sources of disagreement spanning data, task, and annotator factors. We then synthesize modeling approaches using a common framework defined by prediction targets and pooling structure, highlighting a shift from consensus learning toward explicitly modeling disagreement, and toward capturing structured relationships among annotators. We review evaluation metrics for both predictive performance and annotator behavior, noting that most fairness evaluations remain descriptive rather than normative. We conclude by identifying open challenges and future directions, including integrating multiple sources of variation, developing disagreement-aware interpretability frameworks, and grappling with the practical tradeoffs of perspectivist modeling.
\end{abstract}

\section{Introduction}
NLP applications largely rely on supervised learning, which depends on annotated data. Annotation is often operationalized through majority voting, an assumption that can be problematic, particularly for complex tasks where true experts may be outnumbered, or for inherently ambiguous tasks that admit multiple valid interpretations. Majority aggregation can also obscure minoritized perspectives, leading to biased models and representational harm \cite{blodgett-etal-2020-language, 10.1145/3411764.3445423}. This “single ground truth” assumption has been increasingly challenged, especially with the rise of subjective NLP tasks such as toxic language detection and quality estimation. Early critiques appear in \citet{AroyoWelty2015TruthIsALie}, which questions the existence of a unique truth in crowdsourced annotation. This shift aligns with broader efforts to make NLP systems more inclusive, as consensus-based labels often disadvantage minority viewpoints \cite{blodgett-etal-2020-language, 10.1145/3411764.3445423}. Building on this, \citet{Cabitza_2023} formalized \textit{perspectivism} in NLP, advocating for models that integrate diverse human viewpoints rather than collapsing them into gold standards.

We formalize a taxonomy of disagreement by synthesizing prior work into three sources: data, task, and annotator factors. We trace the evolution of NLP approaches for learning from disagreement, from latent-truth models to multi-annotator, embedding-based methods. To unify these developments, we introduce a synthesis table highlighting key trends, including the shift from treating disagreement as noise to modeling it as a prediction target, and the growing emphasis on structured relationships among annotators. We also survey evaluation practices and fairness considerations. Finally, we outline open challenges and future directions, including jointly modeling multiple sources of variation and navigating the practical tradeoffs of perspectivist approaches. 

\begin{figure*}[!tb]
    \centering
    \includegraphics[width=0.8\textwidth]{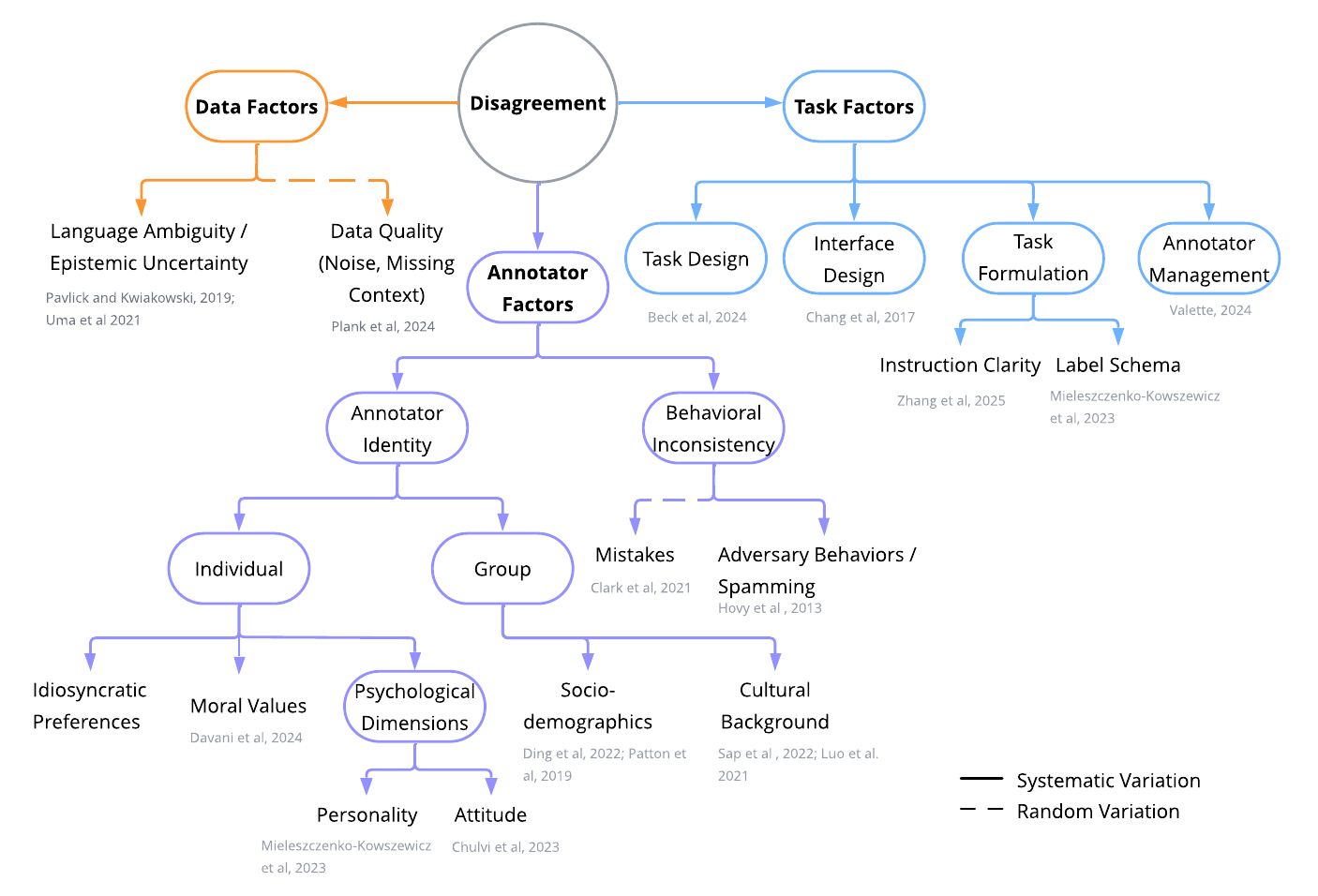}
    \caption{Taxonomy of sources of annotator disagreement, with key citations for each sub-category. For each source of disagreement, we denote systematic variation using solid lines, and random variation using dotted lines.}
    \label{fig:tax}
\end{figure*}

Our distinct contribution is a model-centric, NLP-focused analysis synthesis that does not exist previously: 1) we organize 110+ works into three modeling families and map them onto \textbf{a unified taxonomy of disagreement} across data-, task-, and annotator-driven sources (Fig. \ref{fig:tax}), which reveals that data and task factors are undermodeled; 2) we introduce \textbf{a synthesis framework of disagreement-aware methods}, along prediction  target and pooling structure (Fig. \ref{fig:syn}), which finds that the field is moving toward predicting distributions, and modeling using structured relationships among annotators (partial pooling) is empirically common and the most expressive regime; and 3) we present \textbf{findings on fairness evaluations}, that fairness assessment in this area remains descriptive rather than normative. We discuss survey scope and selection criteria in \ref{sec:app-survey-scope}.


\section{Data}
Research in perspectivist modeling has been constrained by the availability of disaggregated datasets with annotator information. \citet{frenda2024perspectivist} present a comprehensive selection of disaggregated datasets, highlighting that only a few report information about annotator identity. Since that survey, more public data has been released, accelerating progress in modeling (Table \ref{tab:disaggregated-datasets}), spanning multi-modal and multi-lingual contexts.

\section{Related Work}
\subsection{Previous Surveys on Perspectivist NLP}
Two recent surveys are closely related but differ in scope: \citet{frenda2024perspectivist} offers a broad conceptual overview of perspectivist NLP, while \citet{Uma2021LearningFD} emphasizes empirical training and evaluation across vision and NLP. Although \citet{Uma2021LearningFD} also provide a taxonomy, theirs is derived from disagreement observed in six datasets they focus on, most of which involve objective tasks (POS tagging, coreference annotation, medical relation extraction, recognizing textual entailment, and two image classification tasks). Specifically, they explicitly place subjectivity outside their scope. Annotator identity and subjectivity are not included in their taxonomy, but they are the core sources of disagreement in perspectivist literature \cite{Cabitza_2023, frenda2024perspectivist}. On the other hand, our taxonomy decomposes annotator identity into individual and group identity, and more granular ones such as moral values, personality, and sociodemographics, which appear in many NLP annotation datasets. Their taxonomy also has no distinction between systematic and random variation, which would inform whether a source of disagreement is learnable structure or noise. We organize our taxonomy around this structure. Overall, our taxonomy is domain-agnostic and modeling-focused, and exists to motivate and structure our later synthesis of modeling approaches. The taxonomy is just one of our contributions.

\subsection{Previous Taxonomies on Sources of Disagreement}
Prior work has analyzed sources of annotator disagreement across tasks and domains. \citet{basile-etal-2021-need} identify three broad contributors: individual differences (partly linked to sociodemographics), stimulus characteristics (linguistic or task ambiguity), and contextual inconsistency.  \citet{sandri-etal-2023-dont} propose a finer split—sloppy annotations, ambiguity, missing information, and subjectivity—finding subjectivity dominant in offensive language detection. In legal NLP, \citet{xu2024dissonanceinsightsdissectingdisagreements} highlight narrative uncertainty and annotation context. These taxonomies are largely domain-specific, emphasizing different facets depending on task formulation. More importantly, our taxonomy, unlike previous work, is motivated as a foundation for our modeling synthesis (Figs. 2-4), and is organized around sources of disagreement that could be represented during modeling. \citet{basile-etal-2021-need} argues about the evaluation of disagreement, where the sources of disagreement are introduced to motivate that argument rather than to support modeling. \citet{sandri-etal-2023-dont}'s taxonomy is focused on items, and a label schema applied to texts, validated on a single task (offensive language detection). Annotator identity is framed as an inferred property of an item (e.g., a text belongs to a divisive topic, and disagreement occurs due to personal bias), rather than a variable a model can condition on.

\section{Taxonomy of disagreement}
We propose a unifying, domain-agnostic taxonomy (Fig.~\ref{fig:tax}) that synthesizes these views into three sources of disagreement: data factors (e.g., ambiguity and data quality), task factors (e.g., formulation and interface design), and annotator factors (e.g., demographics, preferences, and errors). Different sources of disagreement can interact with one another. Unclear task formulation can lead to inconsistent annotation. Linguistic ambiguity can lead to disagreement that varies with annotators’ individual and group identities, leading to systematically different interpretations of the same text. Our taxonomy is organized around sources of variation during annotation, and each source corresponds to something a disagreement-aware model can represent: annotator factors are decomposed into individual and group levels that need to be represented with different pooling structures that we discuss later in the paper. We also distinguish between systematic and random variation, which determines whether a source is a learnable structure in modeling.

\subsection{Data Factors}
Prior work identifies three data-driven sources of disagreement: data quality issues, linguistic ambiguity, and epistemic uncertainty \cite{Poesio2020Ambiguity, Uma2021LearningFD, PavlickKwiatkowski2019InherentDA}. Missing context or noisy inputs exacerbate inconsistency, especially for short text \cite{Plank2014LearningPW}. Language is inherently ambiguous, and in many subjective tasks disagreement reflects irreducible multiplicity rather than annotation error \cite{PavlickKwiatkowski2019InherentDA}.





\subsection{Annotator Factors}


\noindent \textbf{Individual identities.} Prior work documents stable, annotator-specific biases that introduce variance into training data and directly affect model performance \cite{otterbacher2018social, larimore-etal-2021-reconsidering, PavlickKwiatkowski2019InherentDA, geva-etal-2019-modeling}. Beyond general disagreement, psychological and moral traits systematically correlate with annotation behavior. Personality dimensions such as agreeableness and conscientiousness are associated with evaluative leniency \cite{MieleszczenkoKowszewicz2023CapturingHP}, while moral values exert stronger influence on judgments of offensiveness than geographic or cultural background \cite{10.1145/3630106.3659021}. Together, these findings show that annotators bring stable individual preferences that shape how they interpret and label language.

\noindent \textbf{Group identities.}
Group-level attributes such as gender, race, age, and political orientation influence annotation behavior, particularly in socially grounded tasks. Gender-linked variation has been observed in hate speech detection \cite{Wojatzki2018DoWP} and even in syntactic tasks \cite{garimella-etal-2019-womens}, aligning with sociolinguistic evidence of gendered language use \cite{mondorf2002gender}. In subjective domains such as toxicity or sentiment, demographic shifts can substantially affect model performance \cite{10.1145/3555632}. Qualitative work further shows that lived experience shapes interpretation: community insiders label gang-related language differently from academic annotators \cite{Patton2019Annotating}, and political or racial attitudes correlate with toxicity judgments, including treatment of African American English \cite{sap-etal-2022-annotators, sang2021originvaluedisagreementdata, luo2021detectingstancemediaglobal}. In these settings, disagreement often reflects social perspective rather than noise \cite{Chulvi2023SocialOI}. However, identity effects are task-dependent. For less socially salient tasks (e.g., word similarity, sentiment, NLI), demographic attributes do not consistently explain annotation differences \cite{biester-etal-2022-analyzing}, and explicitly modeling sociodemographics may yield limited gains \cite{orlikowski-etal-2023-ecological}. These findings caution against ecological fallacies \cite{robinson1950ecological}: group-level identity does not uniformly predict individual annotation behavior.

\noindent \textbf{Annotator behavioral inconsistencies.} \citet{abercrombie2025consistencykeydisentanglinglabel} found that annotators give inconsistent responses around 25\% of the time across four different NLP tasks. \citet{clark-etal-2021-thats} also show that untrained annotators identify GPT-3- versus human-written text only at chance levels. Annotators may also produce low-quality or strategically chosen labels to maximize pay \cite{hovy-etal-2013-learning}.

\subsection{Task Factors}
Annotation task design, including its formulation, structure, and presentation, strongly shapes disagreement. Ambiguous or underspecified tasks permit multiple valid interpretations, leading to systematic divergence; in LLM-as-Judge settings, ambiguous prompts elicit different interpretive strategies \citep{zhang2025diverging}. Interface design also affects agreement: unclear guidelines increase inconsistency, while alternative interfaces can surface ambiguity rather than suppress it \citep{Chang2017}. Annotation schemas and scales further modulate disagreement: finer-grained ratings amplify variability relative to binarized labels \citep{MieleszczenkoKowszewicz2023CapturingHP}, and different schemas (e.g., scalar ratings vs. pairwise preferences) capture related but distinct judgments, particularly in RLHF \citep{dsouza-kovatchev-2025-sources}. Presentation and labor conditions matter: order effects, fatigue, and satisficing shift responses over time \citep{Strack1992OrderEffects, Krosnick1996Satisficing, Galesic2009QuestionnaireLength, beck-etal-2024-order}. Fine-grained schemes may inflate disagreement due to degraded consistency under poor working conditions \citep{valette-2024-perspectivism, Yang2023EthicsDataWork}. Data and task factors are not independent: the same text can be unambiguous under one task formulation but ambiguous under another, since task difficulty reflects the interaction between label schema and text properties.

\section{Learning from Disagreement}

We discuss three families of disagreement modeling, tracing NLP’s shift from treating disagreement as noise to modeling it as structured variation. Early work assumed a single gold label, while recent work models disagreement directly via task-based and embedding-based annotator learning.



\subsection{Latent Truth \& Annotator Reliability Modeling}

Models in this family attribute disagreement to annotator reliability, bias, or task difficulty, aiming to infer latent true labels while estimating annotator behavior. They span various contributing factors, capturing task factors (e.g., difficulty), annotator factors (e.g., competence), and approaches that condition annotator behavior on the input. \citet{dawid1979em} introduced a foundational EM-based framework jointly inferring latent labels and annotator error rates from noisy annotations, underpinning many later Bayesian and discriminative models. Later work adds task characteristics, modeling difficulty \cite{whitehill2009vote, ma2015faitcrowd} or expertise-aligned latent topics \cite{fan2015icrowd, ma2015faitcrowd, zhao2015crowdselection, welinder2010multidimensional}, or focuses on annotator competence as scalar accuracies \cite{demartini2012zencrowd, karger2011iterative, liu2012variational}, confusion matrices \cite{dawid1979em, raykar2010learning, liu2012variational}, or bias--variance decompositions \cite{welinder2010multidimensional, raykar2010learning}. Confusion-matrix models are more expressive and typically outperform simpler ones \cite{10.14778/3055540.3055547}. Formally, these models posit a single hidden label $z_i$ per item, treating disagreement as noise around it. They model annotations as $p(y_{ij} \mid z_i, a_j)$, where $a_j$ denotes annotator-specific parameters (e.g., a confusion matrix), and marginalize over $z_i$ at inference. The objective \cite{dawid1979em, hovy-etal-2013-learning} maximizes the marginal likelihood over a single true label per item:
$
\max_{\{a_j\}} \sum_i \log \sum_{z_i} p(z_i) \prod_j p(y_{ij} \mid z_i, a_j).
$
Disagreement thus reflects annotator unreliability rather than item ambiguity.

Later work integrates these ideas into learning architectures. Neural extensions such as CrowdLayer \cite{RodriguesPereira2018Deep} and label-transfer approaches \cite{Tanno2019Learning} replace explicit confusion matrices with differentiable components, while other models introduce worker weighting or structured noise modeling \cite{Gao2022SampleWise, Cao2023Learning, Wei2022Union, Chu2021CommonConfusions}. \citet{Paun2018} show that partial pooling---i.e., drawing annotator parameters from a shared population distribution---best balances expressivity and generalization. Despite their strengths, latent truth models can struggle with scalability and estimation under sparse labeling or weakly differentiated answers. To address this, CROWDLAB \cite{goh2023crowdlabsupervisedlearninginfer} combines annotator statistics with task-level features by treating classifier predictions as an additional annotator, improving robustness without iterative inference. Recent work further incorporates task features and demographic structure: \citet{simpson2015language} jointly models text, annotator bias, and latent truth, while NUTMEG \cite{ivey2025nutmegseparatingsignalnoise} extends truth inference to demographic subpopulations, estimating competence and predicting labels per group rather than collapsing to a single consensus \citep{hovy-etal-2013-learning, Paun2018}.

\subsection{Task-based Annotator Models}

This model family treats each annotator as a distinct task, modeling labeling behavior directly rather than collapsing annotations into a single truth. This family reframes disagreement from noise to a structured signal, and primarily focuses on annotator factors. Given an input $x_i$, a shared encoder produces a
representation $h_i = f(x_i)$, and each annotator $j$ defines a task-specific predictor   $p^{(j)}(y \mid x_i)$. Model parameters are learned by minimizing a supervised loss over
annotator--item pairs, $
\min_{\theta}
\;\;
\sum_i \sum_{j \in \mathcal{A}_i}
\ell\!\left(
y_{ij},\; p^{(j)}_{\theta}(y \mid x_i)
\right)$. $\theta$ is all model parameters (shared encoders and annotator-specific heads), $\mathcal{A}_i$ is set of annotators who labeled $i$, and $\ell(\cdot,\cdot)$ is typically cross-entropy for classification or squared error for regression. Disagreement is captured implicitly through divergence across annotator-specific
predictions. 

Early work by \citet{cohn-specia-2013-modelling} models annotators as correlated tasks in a Gaussian Process framework, using an inter-annotator covariance matrix to control pooling. This nonparametric analogue to hierarchical Bayesian models identifies annotator similarity and outliers, with partial pooling performing best \citep{Paun2018}. Neural extensions implement this idea via shared backbones and annotator-specific heads: CrowdLayer \citep{RodriguesPereira2018Deep} adds a per-annotator layer that learns each labeler's reliability and bias end-to-end through backpropagation, while \citet{Guan2018WhoSaidWhat} model each labeler with a dedicated head and learn sample-specific weights to combine them.  \citet{fornaciari-etal-2021-beyond} further reinterpret disagreement as a regularization signal to reduce overconfidence on ambiguous inputs, learning from both the gold labels and the distribution over multiple annotators (which they
treat as soft label distributions in a single auxiliary task). More recent work models richer annotator structure. Mixture-of-experts approaches capture feature-level heterogeneity \citep{han2025mixtureexpertsbasedmultitask}, while multi-head Transformers preserve uncertainty aligned with empirical disagreement \citep{davani-etal-2022-dealing}. Group-aware and loss-based extensions incorporate demographic structure or explicit tradeoffs between denoising and minority perspectives \citep{fleisig-etal-2023-majority, jinadu2024noisecorrectionsubjectivedatasets}. Collectively, task-based models capture structured perspective variation beyond latent-truth formulations.

\subsection{Embedding-based Annotator Models}
Focusing on data and annotator factors, the third family of embedding-based models departs from latent-truth paradigms by treating disagreement as systematic variation in annotator behavior rather than noise around a hidden true label. Instead of assigning each annotator a separate prediction head, these approaches encode annotator differences in a shared latent space, enabling scalability to thousands of annotators with sparse labels, an important limitation of task-based methods.  Embedding-based models parameterize annotator-level predictions as
$
p(y_{ij} \mid h_i, e_j)
$, where 
 $h_i = f(x_i)$ is an embedding of the input item and
$e_j = g(a_j)$ is an embedding of the annotator $j$.
In practice, the conditional distribution is implemented as $p(y_{ij} \mid h_i, e_j) = \mathrm{Softmax}(\phi(h_i, e_j))$,where $\phi$ is a learned interaction function. The loss function optimizes standard supervised losses over annotator–item pairs, where $\theta$ is all model parameters (shared encoders and annotator-specific heads), $\mathcal{A}_i$ is the set of annotators who labeled $i$, and $\ell(\cdot,\cdot)$ is typically instantiated as cross-entropy for
classification or squared error for regression:
$
\min_{\theta}
\;\;
\sum_i \sum_{j \in \mathcal{A}_i}
\ell\!\left(
y_{ij},\;
p_{\theta}(y \mid h_i, e_j)
\right)
$. Disagreement is formalized as the interaction between item and annotator embeddings.

Early work introduced annotator embeddings to capture individual biases in subjective NLP tasks \citep{9679002}. AART \citep{mokhberian-etal-2024-capturing} extends this idea to large-scale crowdsourcing with regularization for robustness and fairness. Subsequent models enrich the framework by incorporating annotation embeddings \citep{deng-etal-2023-annotate} or demographic structure. Jury Learning \citep{Gordon_2022} combines content, annotator, and demographic embeddings to predict individual votes and compose configurable juries, while DEM-MoE \citep{xu2025modelingannotatordisagreementdemographicaware} explicitly models demographic structure via a mixture-of-experts to capture intersectional variation. Recent work further improves scalability by replacing stored embeddings with hypernetwork-generated, annotator-specific LoRA adapters\citep{ignatev2025hypernetworksperspectivistadaptation}. Embedding-based models show a progression in the granularity of disagreement they represent: from individual annotators \citep{9679002}, to demographic groups \citep{Gordon_2022}, to population-level distributions \citep{weerasooriya-etal-2023-disagreement}. This trajectory points toward an important future direction: estimating subgroup-specific disagreement rather than averaging across populations. Building on prior work \citep{sap-etal-2022-annotators, Lakkaraju2015Bayesian}, future work could focus on inferring how social subgroups interpret the same item, enabling perspective-aware predictions that surface when disagreement reflects broader social divides.

\noindent \textbf{Direct disagreement modeling}. 
A subset of embedding-based models \textit{predicts} disagreement directly rather than treating it as noise. These approaches vary in how disagreement is represented—implicitly via soft supervision, as a scalar score, or explicitly as a label distribution—but share the goal of modeling disagreement as the prediction target. Embedding-based architectures are well-suited to this setting because they enable generalization across annotators, conditioning on demographic attributes, and marginalization over annotator populations. Implicit approaches predict aggregate disagreement without explicitly modeling a distribution. For example, \citet{xu-etal-2024-leveraging} predict empirical label proportions from text using cross-entropy against soft labels, while \citet{Wan2023EveryonesVoice} conditions predictions on annotator demographics and regresses a scalar disagreement score. In contrast, distributional modeling treats disagreement itself as the target, enabling prediction even when not all annotators are observed by directly parameterizing an item-level label distribution: 
$
p(y \mid x_i)
$, where $y$ is the expected distribution of annotators that would be produced by the population, for instance $x_i$. Distributional models typically optimize a composite objective \cite{parappan-henao-2025-learning, weerasooriya-etal-2023-disagreement}, combining supervised alignment with individual annotations and a divergence-based term (e.g., KL or JSD) that aligns predicted and empirical item-level distributions. The first term enforces annotator-level alignment, and the second distributional alignment:
$\min_{\theta} \; \sum_i \sum_{j \in \mathcal{A}_i} \ell\!\left( y_{ij},\, p_{\theta}(y \mid x_i, a_j) \right) + \lambda \sum_i \mathcal{D}\!\left( \tilde{p}_i(y) \,\middle\|\, p_{\theta}(y \mid x_i) \right)$. DisCo \citep{weerasooriya-etal-2023-disagreement} models disagreement at the response, item, and population levels, aggregating over annotators at inference to estimate population-level disagreement. The Learning Subjective Label Distribution framework \cite{parappan-henao-2025-learning} extends this by conditioning distributions on sociodemographic attributes and semantic perspectives, enabling subgroup-specific disagreement estimates. Collectively, these approaches move toward unified representations of individual-, group-, and population-level variation.

\section{Mapping Models to Disagreement Sources}

\begin{figure}[!t]
    \centering
    \includegraphics[width=0.4\textwidth]{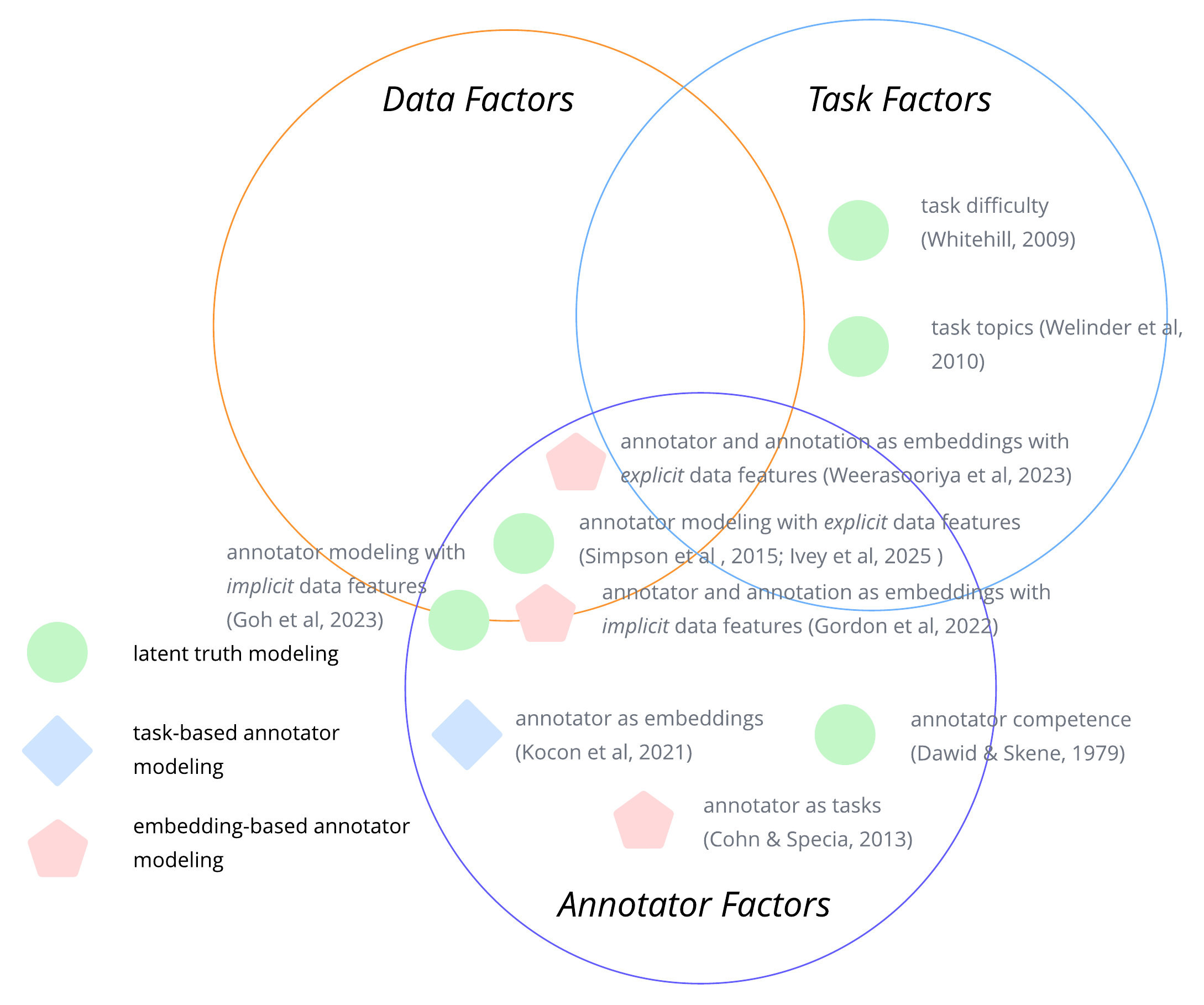}
    \caption{The distribution of the three main methods across our identified taxonomy of sources of disagreement. The lack of work modeling data and task factors point to future directions.}
    \label{fig:venn}
\end{figure}

We map the three model families onto our taxonomy in Figure~2. Latent truth models span multiple regions—task factors, annotator factors, and input-conditioned annotator behavior. Within the annotator–data intersection, we distinguish implicit from explicit data features. Implicit approaches incorporate the input via shared encoders or embeddings without explicitly modeling item ambiguity as a source of disagreement \cite{goh2023crowdlabsupervisedlearninginfer, raykar2010learning, RodriguesPereira2018Deep, Gordon_2022}. CrowdLayer \citep{RodriguesPereira2018Deep} and embedding-based models \citep{Gordon_2022} condition predictions on text but treat it primarily as contextual information. In contrast, explicit data modeling treats properties of the input itself as a structured source of annotator variation \cite{simpson2015language, ivey2025nutmegseparatingsignalnoise}. For instance, \citet{ivey2025nutmegseparatingsignalnoise} condition annotator competence on item features to capture systematic subgroup responses. This category also includes distributional approaches that directly predict item-level disagreement distributions, explicitly representing data ambiguity \cite{weerasooriya-etal-2023-disagreement, parappan-henao-2025-learning}. Notably, the relative scarcity of modeling data factors alone---or jointly integrating data with task and annotator factors---highlights promising directions for future research.

\begin{figure}[!t]
    \centering
    \includegraphics[width=0.49\textwidth]{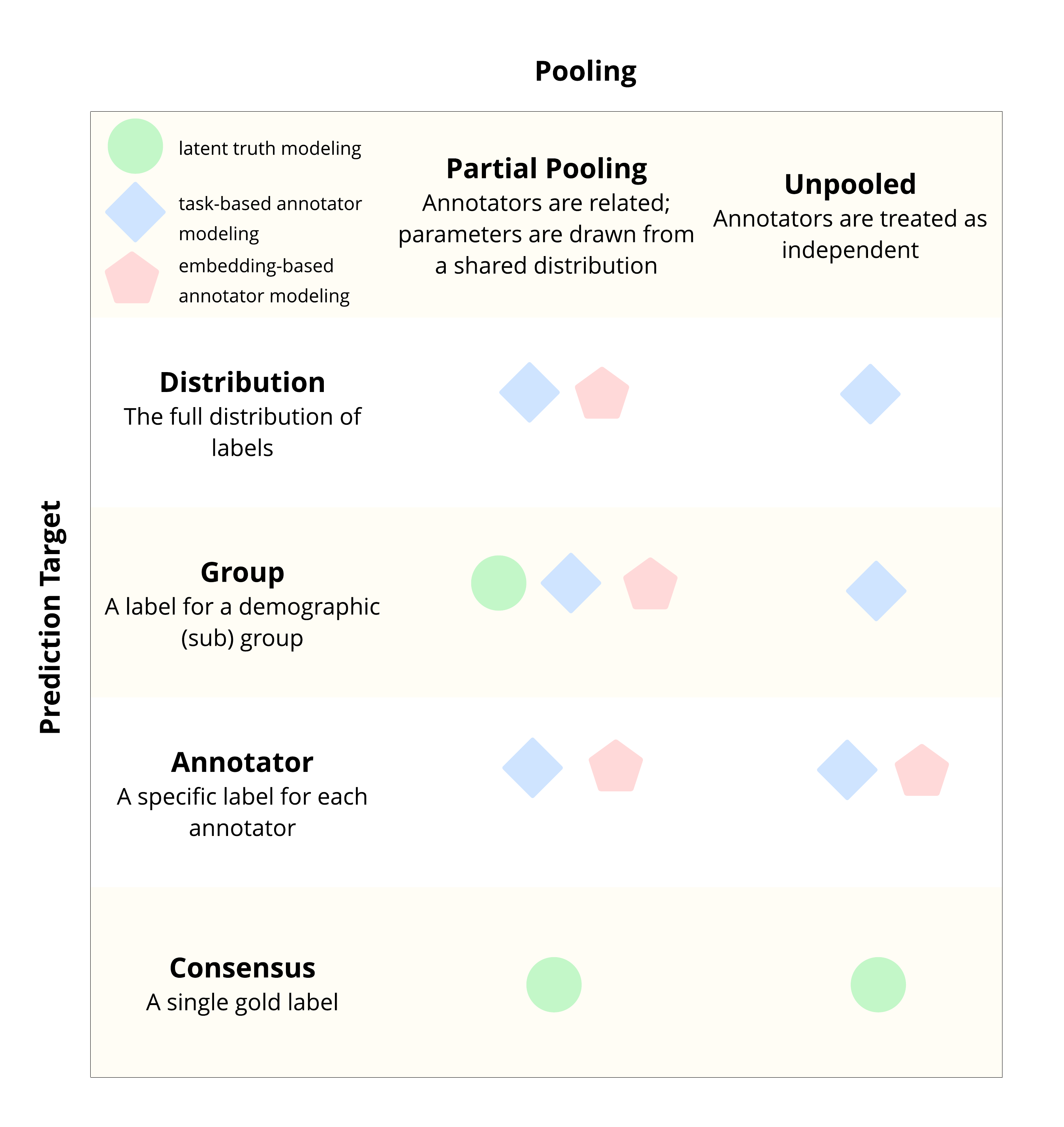}
    \caption{Prediction targets and pooling assumptions across methods for learning from annotator disagreement. The table maps existing approaches by what they predict (consensus labels, individual annotator responses, group-level outputs, or full disagreement distributions) and how they pool information across annotators. Fig. \ref{fig:syn_detailed} is more detailed with representative work.}
    \label{fig:syn}
\end{figure}

\section{Prediction Targets and Pooling}
We introduce a synthesis table (Figure~\ref{fig:syn}) organized along prediction target and pooling structure. Prediction targets include consensus (a single label per item), annotator (one prediction per annotator), group (one prediction per annotator group), and distribution (the full label distribution across annotators). Pooling \cite{Paun2018} is generalized across three model classes. Group pooling forces all annotators to share a single labeling function. Unpooled models treat annotators independently. Partial pooling encodes structured relationships among annotators or between annotators and the (sub)population, though it is instantiated differently across paradigms. In Bayesian latent-truth models, annotator reliabilities are drawn from a shared population prior. Gaussian process multi-task models \cite{cohn-specia-2013-modelling} treat annotators as correlated tasks, while neural multi-task models \cite{RodriguesPereira2018Deep} use shared backbones with annotator-specific heads. Embedding-based approaches \cite{mokhberian-etal-2024-capturing} instead represent annotators as points in a latent population space. Notably, none of the surveyed disagreement-aware models employ full group pooling. While group pooling corresponds to majority vote or fixed aggregation, once annotator behavior is modeled, methods adopt unpooled or partially pooled structures. We highlight these temporal trends:

\noindent\textbf{The field is gradually formalizing disagreement as an object of prediction, not a source of noise}. Modeling has shifted from consensus prediction to individual annotators, to groups, and recently to population or subgroup distributions, an upward movement along the target axis over time.

\noindent\textbf{Across pooling structures, the field has moved from unpooled to partial pooling}. The most successful models use partial pooling (hierarchical priors \cite{Paun2018}, inter-annotator kernels \cite{cohn-specia-2013-modelling}, demographic-informed MoE \cite{xu2025modelingannotatordisagreementdemographicaware}, and subgroup-aware distributions \cite{parappan-henao-2025-learning}) improving generalization and fairness. Partial pooling of annotators or groups is the most expressive and successful regime, and most existing methods fall here, offering generalization to new annotators and scalability under sparse labels.

\noindent\textbf{Few works model disagreement distributions, but newer architectures are converging toward distribution-level modeling} (\citet{fornaciari-etal-2021-beyond} as a task-based model; \citet{parappan-henao-2025-learning, weerasooriya-etal-2023-disagreement} as embedding-based models). Predicting how much people disagree is becoming as important as predicting labels.

\section{Evaluation Metrics}
Evaluation metrics for modeling disagreement broadly fall into two classes: those that compare the predictions to the true annotations, and those that evaluate annotator behavior, aiming to recover latent structure such as inter-annotator relationships.  Hard metrics (accuracy, precision, recall) assume a single gold label, problematic for subjective or ambiguous tasks \cite{poesio-artstein-2005-reliability, Plank2014LearningPW}, motivating a shift toward probabilistic and soft metrics. Aggregation models such as Dawid–Skene \cite{dawid1979em} and MACE \cite{hovy-etal-2013-learning} use likelihood or accuracy against inferred latent truth, while latent-truth and reliability models adopt ROC–AUC, negative log probability, predictive density, and entropy-based measures \cite{raykar2010learning, Paun2018, simpson2015language, ivey2025nutmegseparatingsignalnoise}. As models predict heterogeneous judgments, evaluation extends to per-annotator metrics (MAE, RMSE, macro F1) \cite{cohn-specia-2013-modelling, davani-etal-2022-dealing} and group-level/distributional divergences (KL, JSD) \cite{mokhberian-etal-2024-capturing, Gordon_2022, weerasooriya-etal-2023-disagreement}. However, recent work \cite{rizzi-etal-2024-soft} has shown that Cross Entropy (and to a lesser extent divergence-based measures) may violate desirable properties like symmetry or fair penalization, potentially obscuring meaningful differences between models that aim to reproduce disagreement distributions.


A complementary set of metrics evaluates annotator quality directly. Likelihood-based measures and annotator–model correlations assess inferred behavior \cite{passonneau-carpenter-2014-benefits, Paun2018}, while agreement metrics like Krippendorff’s $\alpha$ and Cohen’s $\kappa$ quantify consistency but conflate ambiguity with unreliability \cite{Krippendorff1980ContentAnalysis, VieraGarrett2005Kappa}. CrowdTruth \cite{Inel2014CrowdTruth} treats disagreement as signal via worker-, item-, and label-level metrics, later extended to model their interactions \cite{dumitrache2018crowdtruth20qualitymetrics}. Recent work targets latent structure more directly, using intra-annotator consistency \cite{10.1145/3411764.3445423} and relational metrics (DIC, BAE) to test whether models preserve agreement geometry \cite{zhang2025unifiedevaluationframeworkmultiannotator}. Overall, evaluation increasingly emphasizes models' preservation of disagreement structure, not merely match labels.

\noindent \textbf{Evaluating Fairness.} Fairness becomes relevant when disagreement is systematic and socially structured, reflecting genuine differences in perspective tied to demographics or lived experience \citep{Aroyo2013CrowdTH, PavlickKwiatkowski2019InherentDA}. Collapsing annotations into a single consensus can erase minority viewpoints, rendering pooling a normative---rather than purely technical---choice. Although fairness is increasingly invoked in disagreement-aware modeling, most evaluations remain descriptive \citep{binns2018fairness}. Disaggregated and group-level metrics reveal disparities \citep{Gordon_2022, xu2025modelingannotatordisagreementdemographicaware}, but do not specify what outcomes should count as fair. Diagnostic frameworks such as AART \cite{mokhberian-etal-2024-capturing} and PERSEVAL \cite{lo-etal-2025-perseval} report parity gaps without explicit normative grounding. These works treat fairness as an empirical outcome rather than as a procedural principle. This gap motivates adapting fairness-in-ML frameworks — such as statistical parity, equalized opportunity, and subgroup fairness \cite{10.1145/3616865} — to disagreement-aware modeling. One concrete operationalization is to evaluate whether the relational structure of disagreement across annotators and groups is preserved, using metrics such as DIC and BAE \cite{zhang2025unifiedevaluationframeworkmultiannotator}, moving toward explicit normative reasoning about what kind of fairness models ought to achieve.

\section{Ethics of Disagreement Modeling}
Demographic-aware disagreement modeling aims at inclusive representation, yet operationalizing demographic identity as input introduces risks, many beginning at data collection \cite{10.1145/3531146.3533226}: individual risks of privacy encroachment and identity misrepresentation, and community risks of reifying social constructs by treating demographics as isolated variables rather than structural, institutional, and relational phenomena \cite{10.1145/3351095.3372826}. Model design choices likewise encode a moral stance \cite{binns2018fairness, heidari2019moral, mehrabi2021survey, 10.1145/3433949}: consensus pooling risks averaging away minority viewpoints, while partial pooling toward demographic groups can reify group boundaries and flatten within-group variation. As noted in Section 2, group identity does not uniformly predict individual behavior \cite{robinson1950ecological}, so models that assume it does risk an ecological fallacy with normative, not just predictive, consequences. More subtly, models meant to amplify minority perspectives can misrepresent them; under coarse demographic conditioning or sparse subgroup data, predictions may reflect stereotypes \cite{Lee2023DissentingVoices, Santurkar2023WhoseOpinions, hu2024quantifyingpersonaeffectllm}. Finally, since current evaluations largely report disparities without specifying what counts as fair \cite{binns2018fairness}, we advocate adapting fairness-in-ML frameworks to the disagreement-aware setting and grounding evaluation in epistemic justice \cite{Fricker2007EpistemicInjustice} to assess whether diverse viewpoints and the structure of disagreement are faithfully preserved.

\section{Challenges and Future Work}

\noindent\textbf{LLM-based simulation of annotator variation.}
LLMs are increasingly used to simulate annotator variation (through persona prompting and reasoning-based approaches) but these methods remain methodologically limited. These approaches are largely unpooled (e.g., personas are simulated independently), highly sensitive to prompts and model choice, and risk amplifying stereotypes \cite{Lee2023DissentingVoices, Durmus2023SubjectiveOpinions, Santurkar2023WhoseOpinions, blodgett-etal-2020-language}. Empirically, persona variables explain only a small fraction of variance \cite{hu2024quantifyingpersonaeffectllm}, and fine-tuning with demographic metadata appears to rely more on annotator-specific signals than generalizable group structure \cite{orlikowski2025demographicsfinetuninglargelanguage}. More broadly, LLM judgments compress human disagreement and rely on opaque priors, raising concerns about reliability and epistemic validity of replacing human annotation labor \cite{Durmus2023SubjectiveOpinions, Cazzaniga2024GenAI}. Recent work has explored whether reasoning can help LLMs capture disagreement, with mixed results. When LLMs reason internally, naive Chain-of-Thought improves disagreement modeling of RLHF LLMs \cite{ni-etal-2026-reasoning}, but long CoT primarily resolves the final answer rather than calibrating label distributions \cite{chen2026decouplingeffectchainofthoughtreasoning}. Even when external human expert reasoning is provided as context, fine-tuning smaller models on the resulting LLM-generated distributions produces inconsistent results \cite{chen2024seeingbigsmallllms}.

\noindent\textbf{Modeling task-induced disagreement.}
While annotator- and data-level variation are increasingly modeled, task-level sources of disagreement remain underexplored. Instruction phrasing, label schema design, presentation order, and interface choices can systematically shape disagreement, yet are rarely represented explicitly. Prior work shows that task design strongly influences annotation outcomes \cite{zhang2025diverging, dsouza-kovatchev-2025-sources}, but most models incorporate task information only indirectly. Future work could embed task schemas directly, enabling models to capture how task framing interacts with annotators and data.

\noindent\textbf{Integrating task, annotator, and data variation.}
Most existing methods focus primarily on annotator bias as the source of disagreement. A few latent-truth models and embedding models \cite{simpson2015language, ivey2025nutmegseparatingsignalnoise} jointly consider annotator and task factors, but task-based approaches typically do not. Developing models that integrate all three sources, task, annotator, and data, remains an open challenge, partly constrained by data availability.


\noindent\textbf{Interpretability of disagreement-aware models.}

\noindent\textbf{Lack of interpretability in disagreement-aware models.} Despite growing interest in modeling annotator disagreement, explainability has received limited systematic attention. Most existing XAI methods in NLP are designed for single-label prediction and focus on why a model produces a particular output, offering little insight into why annotators or groups disagree or where such differences are encoded in the model \citep{danilevsky-etal-2020-survey, 10.1145/3546577, Nauta2023EvaluatingXAI}. Early perspectivist work, such as \citet{mastromattei-etal-2022-change}, takes a step toward disagreement-aware explanations by contrasting predictions across perspective-specific classifiers and highlighting syntactic structures driving divergent stances, but remains domain-specific and largely qualitative. More recent multi-annotator models provide feature- or attention-based explanations (e.g., TAX \cite{cheng2023taxtendencyandassignmentexplainersemantic}, MaDL\cite{herde2023multiannotatordeeplearningprobabilistic}), yet these approaches typically lack validation against annotator behavior and do not assess whether explanations preserve the structure of disagreement.  Recent proposals such as Behavior Alignment Explainability \cite{zhang2025unifiedevaluationframeworkmultiannotator} address this gap by evaluating whether explanations align with real annotator agreement patterns, reframing explainability as a question of relational and epistemic fidelity. Ideally, disagreement-aware models would allow practitioners to trace how perspectives propagate through the model from features, to representations, to experts, to outputs, and to distinguish meaningful perspective variation from noise or bias.

\noindent\textbf{Generalization and practical tradeoffs.}
Annotator disagreement is highly dataset- and domain-specific, and evidence on the benefits of demographic conditioning is mixed \cite{Gordon_2022, orlikowski-etal-2023-ecological}. Models risk overfitting to dataset-specific noise, suggesting a need for cross-dataset and cross-domain training. Perspectivist models introduce tradeoffs: they require richer annotations and greater computation, complicate filtering decisions, and raise normative questions about whose perspectives should be preserved. As a result, adopting disagreement-aware modeling is not merely a technical choice, but a value-laden one balancing fairness, interpretability, efficiency, and annotation labor conditions \cite{valette-2024-perspectivism}.

\section{Conclusion}
We present a unified view of disagreement-aware modeling, treating disagreement as signal rather than noise. Our taxonomy spans data-, task-, and annotator-driven sources, tracing the shift toward multi-annotator and persona-conditioned models. Key gaps remain (limited distributional modeling and descriptive fairness evaluation), motivating future work on integrated, interpretable, and normatively grounded approaches.

\section{Limitations}
We recognize that our work has several limitations. While thorough, decades of work in disagreement make a fully exhaustive survey difficult. Our survey reviews over 110 papers, and our selection reflects methodological relevance to disagreement-aware modeling. As a result, some related approaches, such as LLM simulations of annotator personas, or other methods that engage with human perspectives tangentially rather than as a modeling target, have been excluded. Additionally, the taxonomy we propose is a simplification. In real-world settings, data, task, and annotator factors sometimes cannot be separated cleanly. Future approaches, particularly those that integrate the interaction between various factors, might require extensions to this taxonomy. Lastly, our discussion of open challenges and future directions reflects our interpretations of the most important gaps in the literature. These should not be interpreted as a comprehensive or normative research agenda. Other researchers might prioritize different directions, such as efficiency, model deployment concerns, or annotation practice design to produce meaning disagreement. 

A key limitation of the disagreement-aware literature (and thus our synthesis) is that disagreement is often treated as an observed property of datasets rather than as an outcome of annotation practice design. Observed disagreement may conflate meaningful perspective variation with annotator factors such as confusion and fatigue. This limits cross-paper comparability. Developing and standardizing annotation practices that intentionally elicit meaningful disagreement remains an important direction for future work.

\section*{Acknowledgments}

This work was supported in part by the National Science Foundation under Grant No. IIS-2143529.

\bibliography{custom}

\appendix

\section{Appendix}
\label{sec:appendix}

\subsection{Survey Scope and Selection}
\label{sec:app-survey-scope}
This survey focuses on research in NLP that studies, models, or evaluates annotator disagreement in text-based tasks. We restrict our scope to NLP settings where disagreement is theoretically meaningful, such as toxicity detection, stance analysis, sentiment, and other subjective or socially grounded judgments, rather than domains where disagreement primarily reflects measurement error. Our goal is not to exhaustively catalog all work on annotation variability (of which there is decades of work), but to synthesize lines of research that treat disagreement as a signal relevant to learning, evaluation, or model design, with a slight focus on connecting the most recent work. 

We survey over 110 prior works spanning disagreement sources, modeling methods, and evaluation paradigms. 
We queried venues such as ACL Anthology, Google Scholar, Semantic Scholar, and arXiv. We use keywords including “annotator disagreement”, “learning from crowds”, “multi-annotator learning”, “annotator bias”, “subjective NLP”, “label distribution”, and “perspectivist NLP”. We prioritized publications in major NLP venues (ACL, EMNLP, NAACL, EACL, TACL, Findings of ACL), as well as interdisciplinary venues where annotator modeling is prominent (AAAI, NeurIPS, CHI, FAccT). We emphasized recent work from 2010-2025, focusing on developments after 2018, for two reasons. First, earlier work on learning from disagreement, which was dominated by latent-truth models, is already thoroughly surveyed by \citet{Uma2021LearningFD}, which scope spans both computer vision and NLP. Second, around 2018, the modeling of individual annotators moved toward neural architectures: the per-annotator output heads introduced by CrowdLayer \citep{RodriguesPereira2018Deep} and \citet{Guan2018WhoSaidWhat} became the architectural template that subsequent task-based work builds on, and that later embedding-based methods reformulate to scale to many annotators under sparse labels. This marks a shift from recovering a single ground truth toward preserving individual and group perspectives. 

Table \ref{tab:survey-categories} provides the counts of papers surveyed in each category we focus on. Papers may appear in multiple categories when relevant. We also reviewed 9 papers on large language model–based persona prompting and demographic conditioning; however, because most of this work focuses on simulating personas rather than explicitly modeling annotator structure or disagreement, we discuss these primarily in the Challenges and Future Work section.

\begin{table}[t]
\centering
\small
\begin{tabular}{lr}
\toprule
\textbf{Survey Category} & \textbf{\# Papers} \\
\midrule
Sources of Disagreement & 30 \\
Disaggregated Datasets (Data) & 24 \\
Latent Truth \& Reliability Modeling & 21 \\
Task-Based Multi-Annotator Learning & 8 \\
Embedding-Based Multi-Annotator Learning & 11 \\
Evaluation Metrics & 14 \\
LLM-Based Persona Simulation & 9 \\
\midrule
\textbf{Total Unique Papers Surveyed} & \textbf{117} \\
\bottomrule
\end{tabular}
\caption{Distribution of papers surveyed across categories. Each paper is counted once, in its primary category.}
\label{tab:survey-categories}
\end{table}

We include work that makes a substantive methodological contribution, defined as proposing a learning algorithm, prediction target, or evaluation framework that explicitly accounts for multiple annotators or divergent interpretations. For each of the three modeling families that we identify (latent truth modeling, task-based multi-annotator learning, and embedding-based approaches), we include seminal and highly-cited work \citep[e.g.,][]{dawid1979em, cohn-specia-2013-modelling,9679002} in order to contrast witih representative recent advances. The review was conducted iteratively. We first followed citation chains from seminal works and tracked emerging work from the Workshop on Perspectivist Approaches to NLP\footnote{\url{https://nlperspectives.di.unito.it/}} and related venues. Our taxonomy of disagreement sources was refined in parallel with the literature review.

Taken together, this scope allows us to position our survey as a complement to existing work. Whereas prior surveys are either primarily conceptual \cite{frenda2024perspectivist} or modality-general \cite{Uma2021LearningFD}, we center NLP as a domain where disagreement is deeply tied to task design, interpretation practices, and annotator identities. By jointly examining modeling approaches, evaluation practices, and their fairness implications within a unified framework, we aim to provide a more technically-grounded synthesis of perspectivist modeling that has not yet been articulated in existing surveys. 

\clearpage
\onecolumn
\begingroup
\small
\setlength{\tabcolsep}{4pt}
\renewcommand{\arraystretch}{1.2}
\begin{longtable}{@{}
    >{\raggedright\arraybackslash}p{\dimexpr 0.24\textwidth - 1.5\tabcolsep\relax}
    >{\raggedright\arraybackslash}p{\dimexpr 0.20\textwidth - 1.5\tabcolsep\relax}
    >{\raggedright\arraybackslash}p{\dimexpr 0.16\textwidth - 1.5\tabcolsep\relax}
    >{\raggedright\arraybackslash}p{\dimexpr 0.40\textwidth - 1.5\tabcolsep\relax}@{}}
\caption{Disaggregated datasets reporting annotator identity or sociodemographic metadata.}
\label{tab:disaggregated-datasets} \\
\toprule
\textbf{Dataset} & \textbf{Task / Domain} & \textbf{Language} & \textbf{Annotator metadata reported} \\
\midrule
\endfirsthead

\caption[]{Disaggregated datasets (continued).} \\
\toprule
\textbf{Dataset} & \textbf{Task / Domain} & \textbf{Language} & \textbf{Annotator metadata reported} \\
\midrule
\endhead

\midrule
\multicolumn{4}{r}{\textit{Continued on next page}} \\
\endfoot

\bottomrule
\endlastfoot

Kumar et al. Toxicity-Perspectives \citep{kumar2021designingtoxiccontentclassification} & Toxicity rating on subjective scale & English (Twitter, Reddit, 4chan) & Gender, age, race/ethnicity, sexual orientation, education, political affiliation, religion, parenthood status, prior experience with toxicity \\
\addlinespace
Measuring Hate Speech \citep{sachdeva-etal-2022-measuring} & Hate speech scoring & English & Race, gender, sexuality, age, education, income, political ideology; decision difficulty \& survey bias \\
\addlinespace
ArMIS \citep{almanea-poesio-2022-armis} & Misogyny \& sexism & Arabic & Religious beliefs, gender \\
\addlinespace
Google Specialized Rater Pools \citep{goyal2022toxicitytoxicityexploringimpact} & Toxicity classification & English & Target-identity rater pools: African American, LGBTQ, US control \\
\addlinespace
Annotators with Attitudes \citep{sap-etal-2022-annotators} & Toxic / hate speech & English & Sociodemographics; racial attitudes, free-speech beliefs, conservatism \\
\addlinespace
POPQUORN \citep{pei-jurgens-2023-annotator} & Offensiveness, politeness, QA, rewriting & English & US-representative: sex, age, race, education \\
\addlinespace
DICES \citep{aroyo-etal-2023-dices} & Conversational AI safety & English & Gender, age, race/ethnicity, education \\
\addlinespace
EPIC \citep{frenda-etal-2023-epic} & Irony detection & English (5 varieties) & Age, gender, nationality/variety \\
\addlinespace
VOICED \citep{weerasooriya-etal-2023-vicarious} & Vicarious offense / noise audit & English & Age, gender, race, political leaning \\
\addlinespace
crowd-enVENT \citep{troiano-etal-2023-dimensional} & Emotion + 21 appraisal dimensions over event descriptions & English & Demographics (gender, age, education), Big-Five personality, current emotional state for both authors and reader annotators \\
\addlinespace
NLPositionality \citep{santy-etal-2023-nlpositionality} & Social-acceptability and toxicity judgements & English (global participants) & Country, age, gender, race/ethnicity, education, religion, native language \\
\addlinespace
MultiPICo \citep{casola-etal-2024-multipico} & Irony (short conversations) & Multilingual & Age, gender, ethnicity, country, employment status \\
\addlinespace
D3CODE \citep{davani-etal-2024-d3code} & Offensive language (cross-cultural) & Multiling. (21 countries) & Age, gender, SES, country/region; 6 moral foundations \\
\addlinespace
IndoToxic2024 \citep{susanto2025indotoxic2024demographicallyenricheddatasethate} & Hate speech \& toxicity subtypes & Indonesian & Disability, ethnicity, religion, gender, LGBTQ+, age, education, job status, presidential vote \\
\addlinespace
CREHate \citep{lee-etal-2024-exploring-cross} & Cross-cultural English hate speech & English (AU, GB, SG, US, ZA) & Country of residence (5 Anglophone countries), gender, ethnicity\\
\addlinespace
CSC \citep{jang-frassinelli-2024-generalizable, jang2023intended} & Sarcasm rating (6-point Likert) & English & gender, age, ethnicity, annotator general sarcasm, personality traits\\
\addlinespace
PRISM Alignment Dataset \citep{kirk2024prismalignmentdatasetparticipatory} & LLM preference / value-laden conversations & English (participants from 75 countries) & Age, gender, employment, marital status, education, ethnicity, religion, English proficiency, country of birth/residence, LLM familiarity \\
\addlinespace
A Perspectivist Corpus of Numbers in Social Judgment\citep{may-etal-2024-perspectivist} & Social/moral judgments with numerical reasoning & English & Gender identity, nationality, age, religion, political orientation, education \\
\addlinespace
GBV dataset \citep{abercrombie-etal-2024-revisiting} & Online Gender-Based Violence & English & Gender, age, ethnicity, social psychology surveys\\
\addlinespace
iNews \citep{hu-collier-2025-inews} & Affective response to news & Multilingual & Age, gender, income, education; Big Five, PANAS, CRT; news trust, consumption, politics \\
\addlinespace
LeWiDi-2025 ``Par'' \citep{leonardelli-etal-2025-lewidi} & Paraphrase detection & English (Quora Question Pairs) & Gender, age, nationality, education \\
\addlinespace
VariErrNLI \citep{sarumi2026finegrainedperspectivesmodelingexplanations} & NLI labels + label-explanation validity judgments & English & Gender, age, nationality (CN, DE), education \\
\addlinespace
Community Alignment Dataset \citep{zhang2026cultivatingpluralismalgorithmicmonoculture} & Multi-turn LLM preference with negatively-correlated candidate sampling & English, French, Italian, Hindi, Portuguese & Country, language, plus age, gender, education\\
\addlinespace
MFRC \citep{trager2026moralfoundationsredditcorpus} & Moral sentiment (Care, Equality, Proportionality, Authority, Loyalty, Purity, Thin Morality) & English & gender, sexual orientation, age, household income, first language, political ideology along a liberal-conservative scale, religious affiliation, and moral values via the Moral Foundation Questionnaire \\
\caption{Disaggregated datasets reporting annotator identity or sociodemographic metadata.}
\end{longtable}

\onecolumn 
\raggedbottom 
\subsection{Detailed Synthesis Table of Prediction Targets and Pooling Structures}

We provide a detailed table mapping representative work in each modeling family to the axis of prediction target and pooling structures (Fig.~\ref{fig:syn_detailed}).

\begin{figure}[H]
    \centering
    \includegraphics[width=0.7\textwidth,height=0.65\textheight]{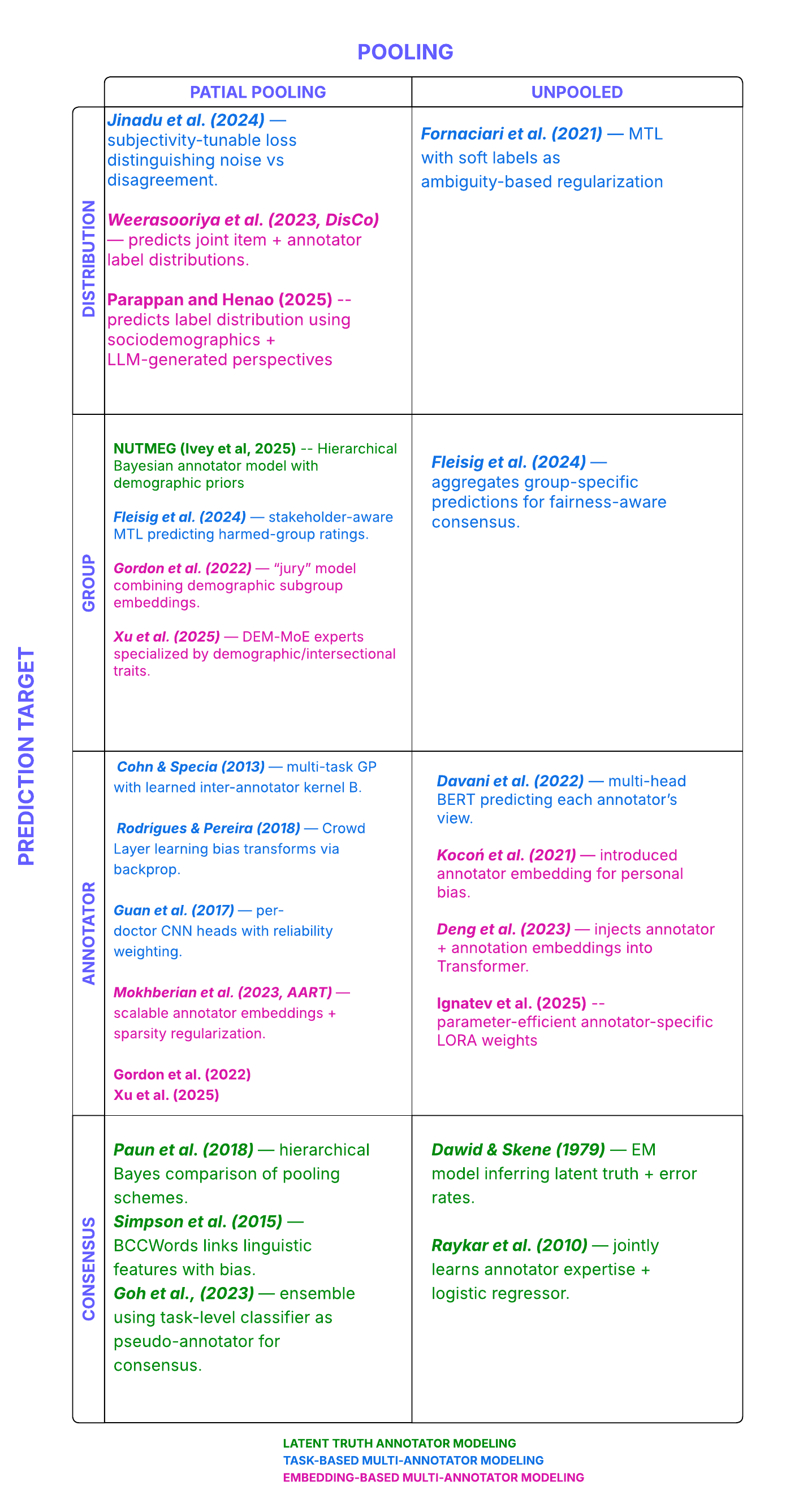}
    \caption{Prediction targets and pooling assumptions across methods for learning from annotator disagreement. The table maps existing approaches by what they predict (consensus labels, individual annotator responses, group-level outputs, or full disagreement distributions) and how they pool information across annotators. Modeling choices implicitly encode different assumptions about population structure, perspective aggregation with recent work increasingly favoring partial pooling and distributional targets.}
    \label{fig:syn_detailed}
\end{figure}

\end{document}